# CROSS-DATASET HARMONIZATION FOR ROBUST ENDOSCOPIC IMAGE ANALYSIS

*Romil Imtiaz, Dimitris K. Iakovidis*

Department of Computer Science and Biomedical Informatics, University of Thessaly, Lamia, Greece

## ABSTRACT

A significant problem in endoscopic image analysis is that the machine learning (ML) models used for this purpose usually underperform when applied on images acquired from endoscopes that are different from those used to acquire the images of their training set. The main difference of the images originating from different endoscopes is their color distributions, which depend both on the image sensors and the light sources used. Although previous studies have highlighted this challenge, to the best of our knowledge it has not been previously explicitly tackled. This study focuses on this problem and proposes very simple but impactful method. It implements a reference-based image harmonization that reduces global appearance differences between endoscopic datasets. Specifically, it extracts global color statistics from a chosen reference dataset in the CIE-Lab color space and applies a statistical channel-wise transformation to map each target image toward the appearance of the images of the reference dataset. The method is evaluated in the context of polyp detection in both flexible colonoscopy and capsule endoscopy datasets using a dataset-level cross validation protocol. The results indicate that the proposed harmonization consistently improves cross-dataset performance up to 30.7%, outperforming relevant baseline and state-of-the-art methods. The results indicate that a substantial part of the generalization gap is driven by low-level appearance variation that can be mitigated without retraining.

## 1. INTRODUCTION

Endoscopic image analysis plays an important role in colorectal cancer prevention, where early detection of lesions such as polyps can support timely intervention. Despite strong progress in automated polyp analysis, machine learning models often underperform when applied to images acquired with endoscope systems different from those used for training. A major cause of this degradation is the shift in global appearance, especially color distributions, caused by differences in image sensors and illumination. Clinical evidence shows that even standard enhancement settings of endoscopy systems can alter AI performance consistency, indicating that image acquisition and processing differences can create a domain shift that could result in misinterpretation of the medical images in practice [1]. Cross-dataset performance degradation has also been reported in endoscopy under realistic transfer settings, where models trained on one source fail to generalize reliably to another source acquired under different conditions [2].

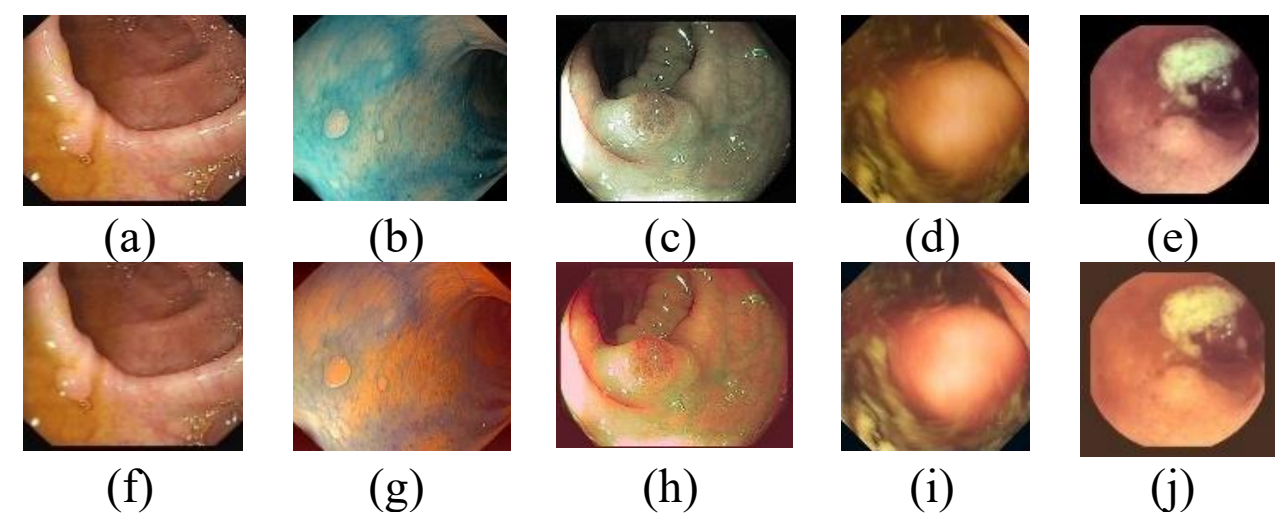

**Fig. 1**. (a–e) Representative samples from five different endoscopy datasets. (f–j) The samples after the application of the proposed image harmonization method, using image (a) as a reference image.

This work targets this device-driven appearance shift using a simple, reference-based image harmonization step. Global color statistics are extracted from a chosen reference dataset in the CIE-Lab color space, and each target image is mapped toward the reference appearance using a lightweight channel-wise statistical transformation. While global-statistics color transfer has been investigated years ago in general vision [3], its utility in the context of harmonization of different endoscopic datasets acquired from different devices has not been previously studied. Most relevant previous works have addressed single-image enhancements, such as linear or non-linear color transformations [4], without considering aligning the color distributions of images acquired using different endoscopic devices. The method presented in this paper does not require retraining of a machine learning-based image analysis system to perform well on images acquired by a different endoscope from that used for the system's training. Sample images from a reference set of images acquired with the same device as the system's training set, are used to compute fixed CIE-Lab statistics for a deterministic preprocessing step. The system can then be applied across datasets with different color distributions using the same trained weights and inference settings. An example is provided in Fig. 1, which demonstrates how the proposed method impacts the appearance of images acquired with different endoscopes, after harmonizing them using Fig. 1(a) as a reference.

Therefore, contribution of this paper is primarily a controlled endoscopy-specific preprocessing protocol rather than a training-heavy adaptation framework. To the best of our knowledge there is no other state-of-the-art method explicitly tackling this problem.

The rest of this paper consists of four sections. Section 2 reviews the state-of-the-art works related to this study, and section 3 details the proposed methodology. The experimental framework and the results obtained are presented in section 4, and the conclusions of this study are summarized in section 5.

## 2. RELATED WORK

Public endoscopy datasets have enabled rapid progress, yet

Author version of the accepted paper for publication in the 2026 IEEE International Conference on Image Processing (ICIP), CCIW Workshop.

heterogeneity in acquisition devices and pipelines makes cross-dataset evaluation and dataset pooling difficult. A review of colonoscopy imaging datasets highlights large variation in availability, accessibility, and usability, as well as in notes that capture details are not consistently reported, complicating principled dataset combination [5]. Representative resources include capsule endoscopy datasets such as Kvasir-Capsule [6] and KID2 [7], while large-scale flexible colonoscopy video data, such as LDPolypVideo [8] further supports robust model development. Challenging flexible colonoscopy datasets, such as EDD2020 [9], continue to emphasize cross-dataset robustness by exposing performance drops under heterogeneous acquisition conditions.

The endoscopic appearance of tissues can vary significantly between different endoscopic systems, mainly due to the different image sensors and illumination sources. Several methods have been proposed to enhance the appearance of endoscopic images; however, they mainly focus on improving image color, contrast and tissue details. Such methods include fusion-based contrast and illumination enhancement [10], texture and color enhancement imaging (TXI) [11], color image reproduction techniques [12], and color transformations for video endoscopic images [4]. However, since these methods are applicable per image, aligning the appearance of tissues between different devices remains a challenge.

To achieve such an alignment application of color normalization/standardization and color-transfer methods has been investigated. For example, state-of-the art foundational models have been developed after preprocessing images from different endoscopic devices using the mean–standard deviation (Z-score) normalization method to reduce intensity-scale differences [13]. Histogram matching and histogram-based standardization have been widely used to align image intensity or color distributions across acquisition settings [14]. More advanced training-free appearance-transfer methods include Fourier Domain Adaptation (FDA), which aligns low-level appearance by replacing low-frequency Fourier amplitude components while preserving structural phase information [15], and Principal Component Analysis/Singular Value Decomposition (PCA/SVD)-based color transfer, which aligns dominant color components and covariance structure between source and reference images [16]. Recent color-transfer studies also describe palette-based transfer as a representative strategy for mapping color groups between images while preserving image content [17].

Most prior work addressing domain shift in endoscopy focuses on preprocessing methods applied before training deep learning models. For example, domain-adaptive frameworks for cross-device colonoscopy explicitly introduce adaptation mechanisms and additional training design to reduce the domain gap [18]. Synthetic data augmentation using generative models has also been explored to expand the training distribution and improve detector generalization [19].

Unlike previous deep learning-based methodologies, the proposed one does not require retraining. Furthermore, methods such as FDA, PCA/SVD, and palette-based transfer require manual choices, e.g., the spectral region size, transformation strength, color space, or the palette size, whereas the proposed method uses fixed CIE-Lab statistics computed from a reference dataset and applies the same deterministic transformation to all target images. The details of the proposed methodology are provided in the next section.

## 3. METHODOLOGY

The proposed method applies to a lightweight dataset harmonization step to reduce appearance differences between endoscopy datasets acquired with different devices. A reference dataset defines the target appearance space. Global color statistics are computed from the reference dataset in the CIE Lab color space, and each target image is transformed to match these statistics, producing a harmonized target set that is evaluated alongside the original target images. This design isolates the effect of appearance alignment, since model architectures, training procedures, and learned parameters remain unchanged. Therefore, any performance change can be attributed to the input appearance shift rather than retraining or parameter tuning.

### 3.1. Problem Setting and Harmonization Objective

Endoscopic image classification models often degrade when evaluated on images acquired with endoscope systems different from those used for training. A key driver of this degradation is device- and pipeline-dependent appearance shift, which is frequently expressed as changes in global illumination and color rendering (*e.g.,* sensor response, light spectrum, optics, and in-camera processing).

To address this effect, the proposed method introduces a reference-based dataset harmonization step that maps a target dataset $D_t$ to the appearance space of a source (reference) dataset $D_s$. Given a fixed reference $D_s$, the method constructs a deterministic transformation $T(.)$ and produces a harmonized dataset $T(D_t \rightarrow D_s)$. The transformation does not require model retraining or parameter optimization and can be applied offline as preprocessing, enabling a controlled evaluation where the only change between raw and harmonized inputs is the input appearance, while the learning pipeline (architecture, training and parameters) remains unchanged.

### 3.2. Image Harmonization

The harmonization step is implemented as a deterministic transformation in CIE-Lab color space, which separates luminance from chromatic components. Furthermore, this color space has been proved effective for endoscopic image analysis, mainly due to its capacity to isolate well red and yellow hues, which are dominant in abnormalities [20]. For a given experiment, the training dataset $\mathcal{D}_s$ is treated as the reference. All images in $\mathcal{D}_s$ are converted from RGB to CIE-Lab, and channel-wise reference mean and reference standard deviation are computed once for each channel $c \in \{L, a, b\}$ using pixels aggregated across the full reference dataset. These reference statistics remain fixed during the experiment.

For each target dataset $\mathcal{D}_t$, images are processed independently. Here, $D_s$ denotes the source dataset, $I \in D_t$ is a target image; $I^{Lab}$ is its CIE-Lab representation; $I_c'^{Lab}$ is the harmonized output, and $I_c^{Lab}$ denotes the $c$-th CIE-Lab channel, where $c \in \{L, a, b\}$. The reference statistics $\mu_c^s$ and $\sigma_c^s$ are computed over pixels of images $I$. Each target image $I$ is converted to Lab and per-image statistics $\mu_c^I$ and $\sigma_c^I$ are computed for each channel. The code then applies a channel-wise affine mapping that aligns the target image's Lab distribution to the reference distribution:

$$I_c'^{Lab} = \left(\frac{\sigma_c^s}{\sigma_c^I}\right)\left(I_c^{Lab} - \mu_c^I\right) + \mu_c^s, c \in \{L, a, b\} \qquad (1)$$

This operation first normalizes the target channel by subtracting its mean and scaling by its standard deviation and then restores it using the reference dataset statistics. A small constant $\boldsymbol{\varepsilon}= 10^{-6}$ is added to standard deviation terms during reference-statistics computation and channel normalization to avoid numerical instability and division by zero when $\sigma_c^I$ is very small. After transformation, values are clipped to valid CIE-Lab ranges and converted back to RGB. The transformed image is saved using the original naming convention to preserve compatibility with existing label files and evaluation scripts. Repeating this process for all images in $D_t$ yields the harmonized dataset $T(\mathcal{D}_t \rightarrow \mathcal{D}_s)$.

This is a simple transformation, but it directly targets a common and well-known source of cross-dataset variation in endoscopy. The complete harmonization procedure is summarized in Algorithm 1. This algorithm employs color transfer as a controlled preprocessing protocol, keeping the detector weights, inference settings, and evaluation pipelines fixed to isolate the effect of appearance. The remainder of the pipeline keeps model architectures and training unchanged and uses the harmonized datasets only as alternative inputs for cross-dataset testing. This design isolates the effect of harmonization, since any performance change can be attributed to the input appearance shift rather than retraining or parameter tuning. In addition, the method can be applied offline as a preprocessing step and does not add inference-time cost to the deployed model.

**Algorithm 1** Dataset Harmonization

**Input:** reference dataset $D_s$, target dataset $D_t$
**Output:** harmonized dataset T ($D_t \rightarrow D_s$)
1: Convert all images in $D_s$to CIE-Lab
2: Compute reference $\mu_c^s$, $\sigma_c^s$ over all pixels in $D_s$ for $c \in \{L, a, b\}$
3: For each image I in $D_t$:
4: Convert I to CIE-Lab
5: Compute per-image $\mu_c^I$, $\sigma_c^I$ for $c \in \{L, a, b\}$
6: Apply Eq. (1) channel-wise to obtain harmonized image I′
7: Convert to RGB, Save image I′.

## 4. RESULTS AND DISCUSSION

We applied a cross-dataset evaluation protocol in which models were trained once on a reference dataset and then tested on independent target datasets acquired under different imaging conditions. The proposed reference-based harmonization was compared with both preprocessing and training-free appearance-transfer methods that have been used in recent relevant works. These include Z-score (mean-variance) normalization in the RGB color space [13], histogram matching [14], Fourier Domain Adaptation (FDA) [15], Principal Component Analysis/Singular Value Decomposition (PCA/SVD)-based color transfer [16], and palette-based color transfer (PBCT) [17]. In addition, Z-score-based variants in alternative color spaces, namely HSV, YCbCr, Lab-L, and Lab-ab [21], were evaluated as lightweight ablation baselines. Z-score normalization was applied to all channels in the HSV and YCbCr color spaces, whereas the Lab-L and Lab-ab variants normalized only the luminance (L) channel and chromatic (a,b) channels, respectively. For FDA, PCA/SVD, and PBCT, the best-performing tested parameter configuration was used, whereas the proposed method directly computes fixed reference-domain statistics and does not require method-specific parameter tuning. For all methods, the trained weights, inference settings, decision threshold, and evaluation pipeline were kept unchanged. Therefore, any performance difference can be attributed to the input appearance transformation rather than retraining, architectural modification, or parameter tuning. Dataset characteristics and acquisition information are summarized in Table I. Experiments were organized into two modality-consistent tracks: flexible colonoscopy, using DS1 (LDPolyp) as the reference dataset and DS2 (EDD2020) and DS3 (KUMC) as target datasets, and capsule endoscopy, using DS4 (Kvasir-Capsule) as the reference dataset and DS5 (KID) as the target dataset. The target datasets contained 127 polyp and 53 normal images (DS2), 103 polyp and 30 normal images (DS3), and 44 polyp and 35 normal images (DS5).

Performance was evaluated using sensitivity, specificity, F1-score (threshold = 0.5), and the area under the receiver operating characteristic (AUC). AUC was emphasized because it reflects probability- ranking ability and is comparatively less sensitive to class imbalance than the aforementioned metrics, which are threshold-dependent [23]. YOLOv7 [24] was used as a real-time detection backbone for polyp localization, while ResNet50 [25] and InceptionV3 [26] were used as classification backbones to verify whether the harmonization benefits were consistent across different model architectures. For image-level evaluation, an image was considered positive if at least one detection corresponding to the polyp class remained after non-maximum suppression (NMS). The maximum detection confidence score was used as the image-level prediction score.

**Table I.** Dataset summary and acquisition devices.

| ID | Dataset | Modality | Data | Labels | Device |
|---|---|---|---|---|---|
| DS1 | LDPolyp [8] | Flexible | Video, Images | Boxes | Olympus |
| DS2 | EDD20 [9] | Flexible | Images | Boxes | Multi-source |
| DS3 | KUMC [22] | Flexible | Images | Boxes | Multi-source |
| DS4 | Kvasir-Capsule [6] | Capsule | Video, Images | Boxes | Olympus EC-S10 |
| DS5 | KID [7] | Capsule | Images | Masks | MiroCam |

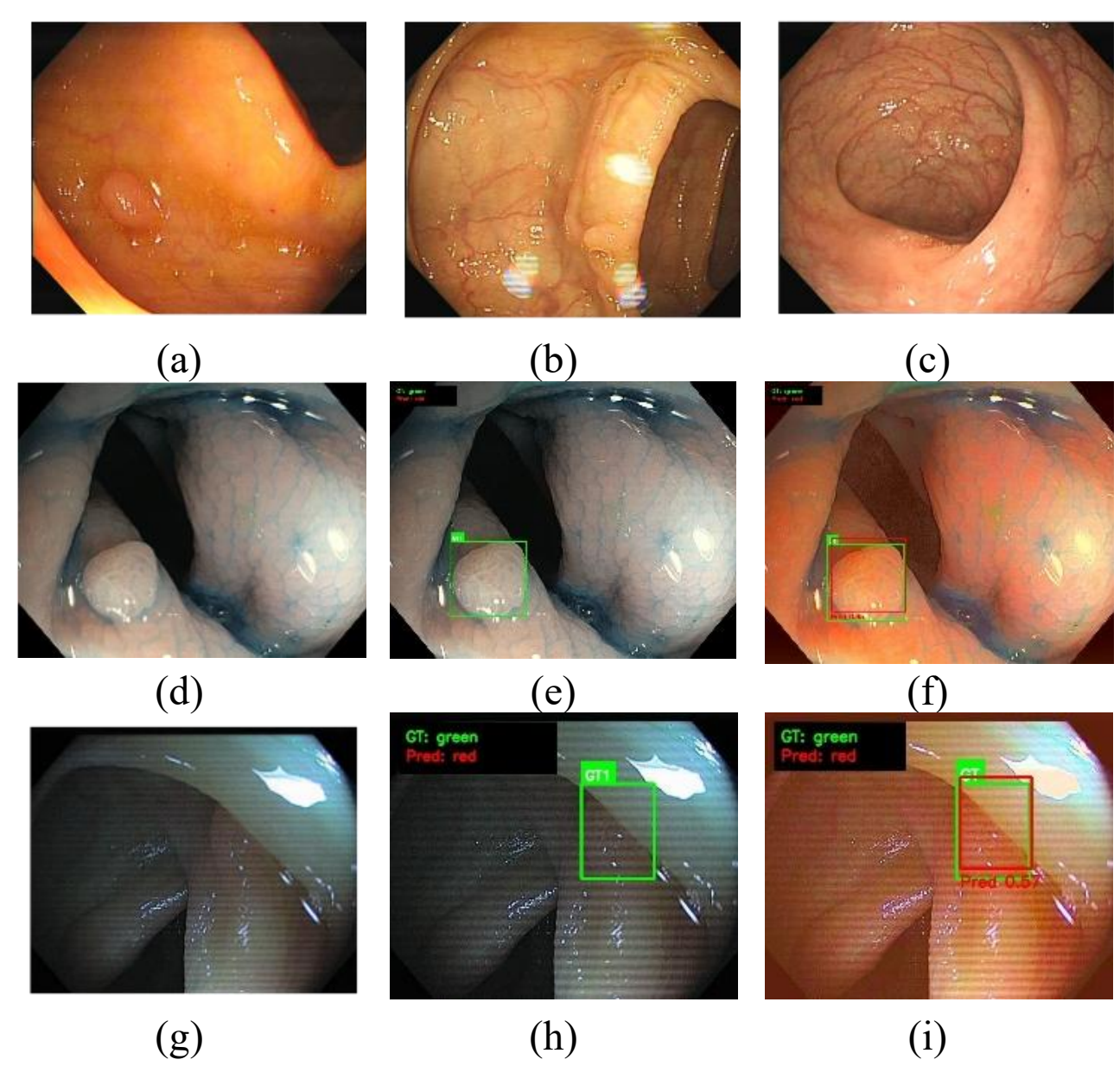

**Fig. 2**. (a–c) Reference images from DS1. (d–g) Raw images without transformation from DS2 and DS3. (e–h) Failed polyp detection in the raw images. (f–i) Successful polyp detection in harmonized images.

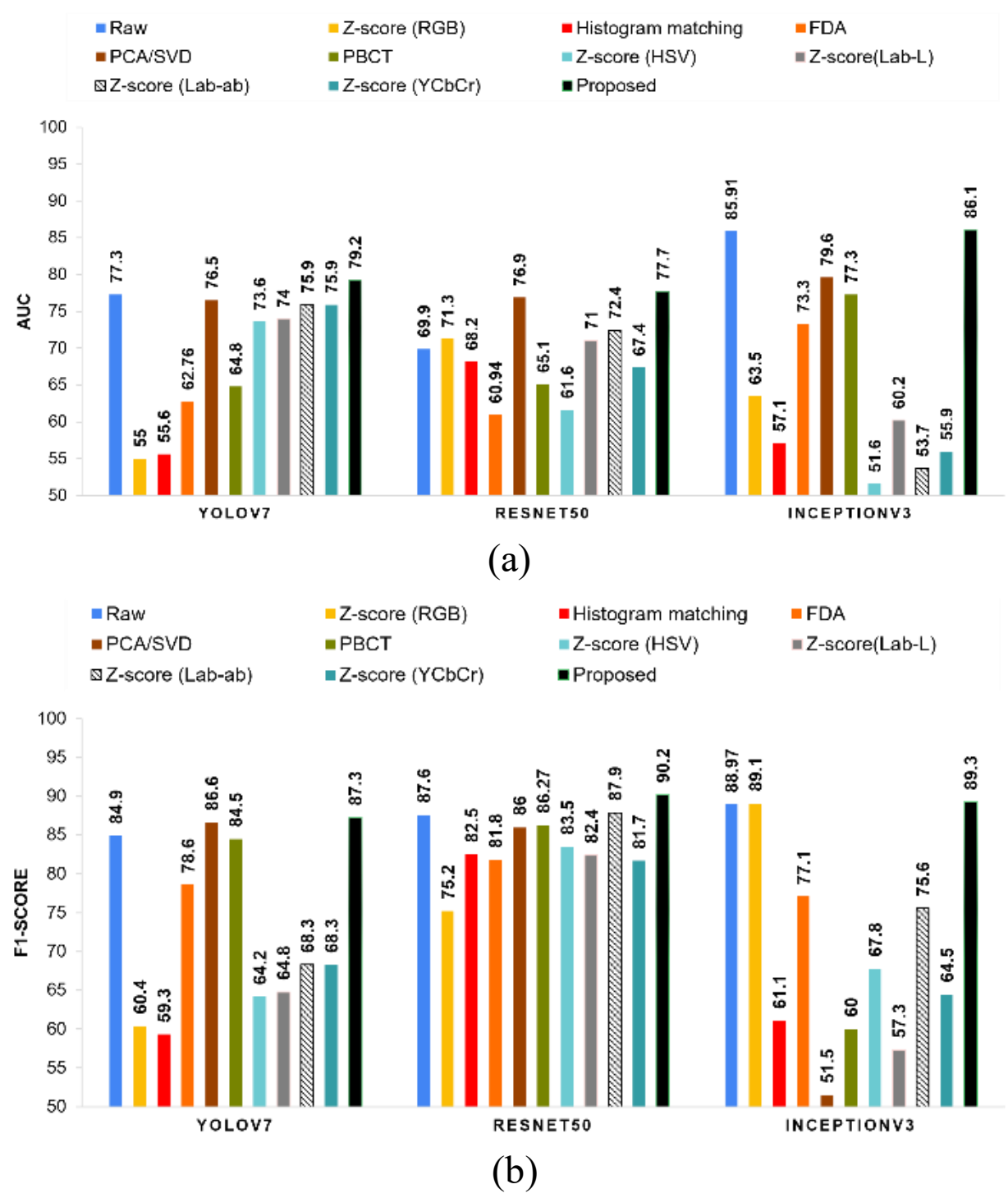

**Fig. 3**. Performance comparison on DS2 using DS1 as the reference dataset for YOLOv7, ResNet50, and InceptionV3. (a) AUC. (b) F1-score.

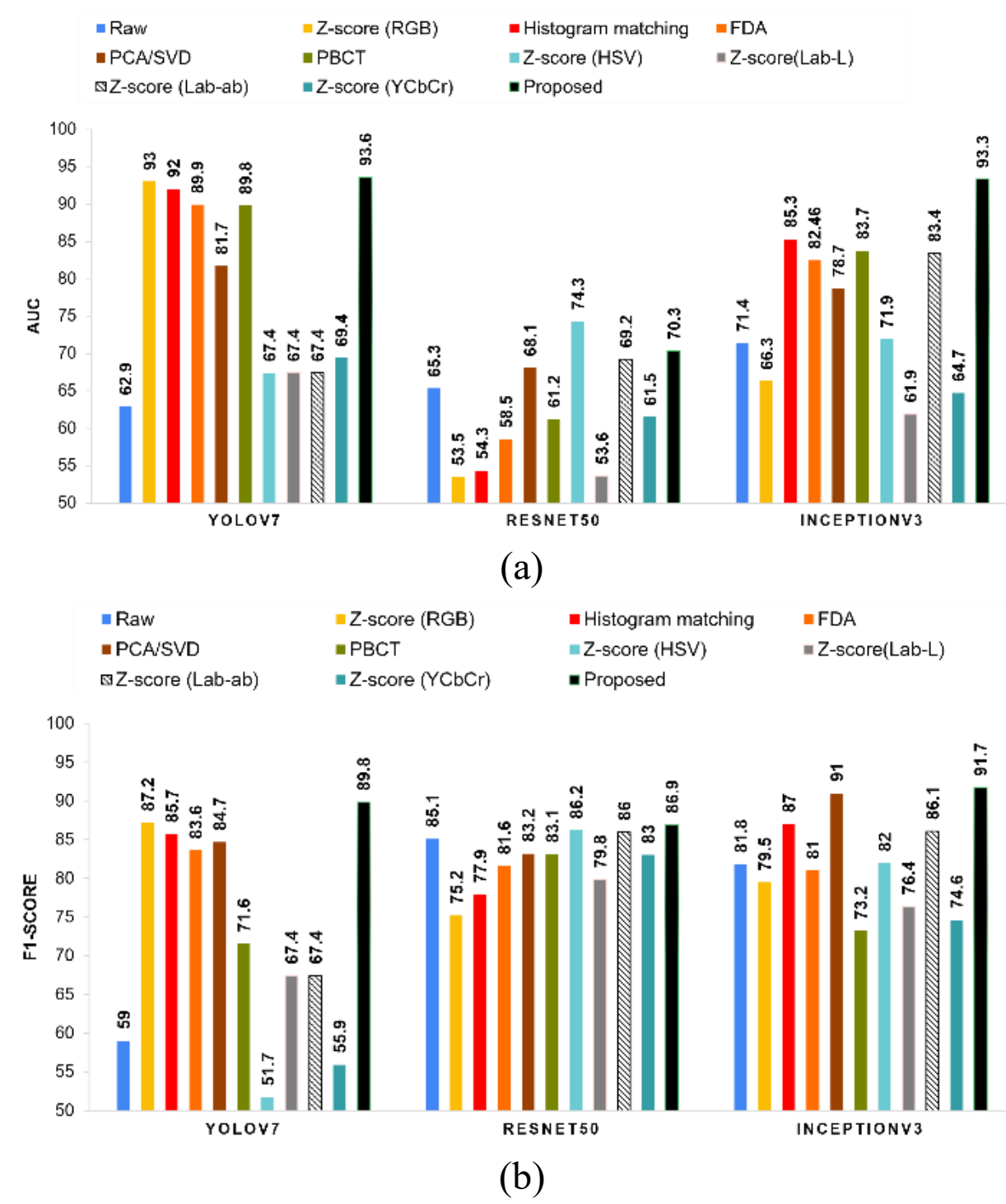

**Fig. 4**. Performance comparison on DS3 using DS1 as the reference dataset for YOLOv7, ResNet50, and InceptionV3. (a) AUC. (b) F1-score.

### 4.1. Flexible colonoscopy

The flexible colonoscopy evaluation is presented using both qualitative and quantitative evidence. Fig. 2 shows representative qualitative examples, including DS1 reference images, raw target images from DS2 and DS3, failed detections on raw images, and successful detections after the proposed harmonization. The quantitative results are presented in Figs. 3 and 4, where DS1 was used as the reference dataset and DS2 and DS3 were used as target datasets, respectively.

Figure 3 shows that on DS2 the proposed harmonization gives the most stable improvement across AUC and F1-score for all three backbones. For YOLOv7, the proposed method improved AUC from 77.3% to 79.2% (+2.4%) and F1-score from 84.9% to 87.3% (+2.7), outperforming FDA and PBCT while remaining close to PCA/SVD in F1-score.

For ResNet50, the proposed method achieved the best overall result, increasing AUC to 77.7% and F1-score to 90.2%, which is higher than all compared state-of-the-art appearance-transfer methods. For InceptionV3, the proposed method also achieved the strongest performance, with AUC increasing to 86.1% and F1-score reaching 89.3%. Although some comparison methods improved one metric in selected cases, their performance was not consistent across all backbones. In contrast, the proposed method improved or preserved both AUC and F1-score across YOLOv7, ResNet50, and InceptionV3, showing better cross-dataset stability.

The sensitivity and specificity results further support this trend. For YOLOv7, the proposed method improved sensitivity from 77.9% to 81.8% (+5%) while maintaining specificity at 76.6%. For ResNet50, sensitivity increased from 86.6% to 91.3% (+5.4%), with specificity remaining stable at 53.3%.

The largest operating-point improvement was observed for InceptionV3, where sensitivity increased from 95.2% to 96.0% (+0.8%) and specificity improved substantially from 20.0% to 95.2% (+376%). This indicates that the proposed harmonization not only improved ranking and F1-score but also produced a more reliable balance between lesion detection and false-positive control.

Figure 4 shows that on DS3 the proposed harmonization achieved the strongest and most consistent performance across the three backbones. For YOLOv7, the proposed method reached the highest AUC of 93.6% and the highest F1-score of 89.8%, outperforming FDA, PCA/SVD, PBCT, Z-score normalization, and histogram matching. For ResNet50, the proposed method also achieved the best AUC and F1-score, improving AUC from 65.3% on raw images to 70.3% (+7.7%) and F1-score from 85.1% to 86.9% (+2%). For InceptionV3, the proposed method produced the clearest gain, increasing AUC from 71.4% to 93.3% (+30.7) and F1-score from 81.8% to 91.7% (+12.1%), which was higher than all comparison methods. Overall, the DS3 results show that the proposed harmonization is not only competitive with state-of-the-art appearance-transfer methods, but also more stable across model architectures and evaluation metrics.

The sensitivity and specificity results further show that the improvement was not only due to AUC. For YOLOv7, the proposed method improved sensitivity from 42.7% to 63.1% (47.7%), showing better lesion detection. For InceptionV3, specificity increased from 20.0% to 50.0% (+150%) while sensitivity increased to 97.0%, indicating a better balance between detecting lesions and controlling false positives.

### 4.2. Capsule endoscopy

The capsule endoscopy evaluation uses DS4 as the reference dataset and DS5 as the target dataset to assess whether the proposed harmonization improves cross-dataset robustness beyond flexible colonoscopy, under challenging capsule-specific variations such as illumination, motion, viewpoint, and gastrointestinal content.

On DS5, the proposed harmonization showed the strongest and most consistent AUC performance across all three backbones, as shown in Fig. 5. For YOLOv7, AUC increased from 75.6% in the raw setting to 85.13% (+12.6%) after harmonization. For ResNet50, the proposed method achieved the highest AUC of 71.3%, outperforming raw images and all comparison baselines. For InceptionV3, the proposed method also achieved the best AUC, increasing from 79.2% to 85.1% (+7.4%).

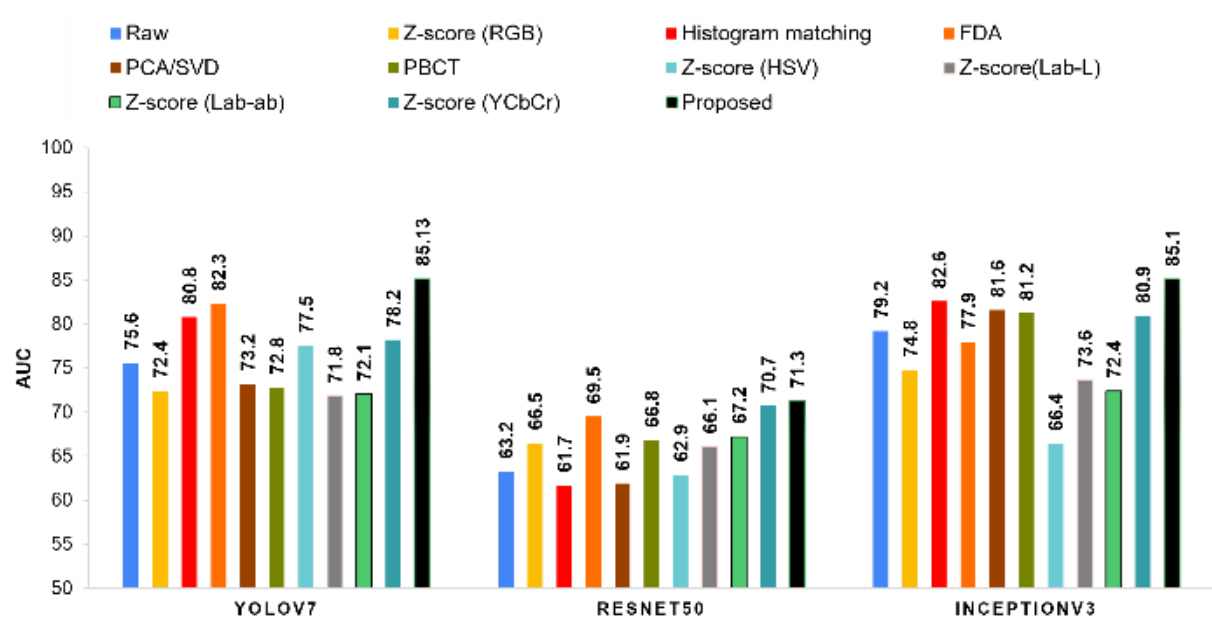

(a)

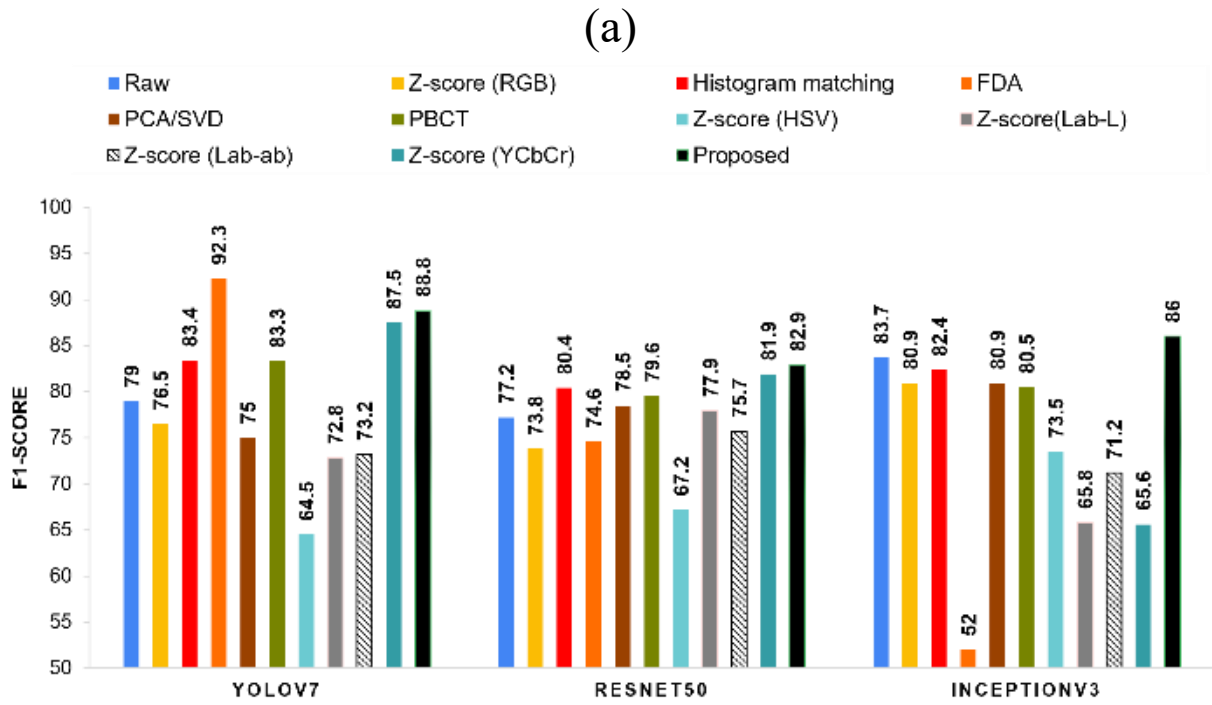

(b)

**Fig. 5**. Performance comparison on DS5 using DS4 as the reference dataset for YOLOv7, ResNet50, and InceptionV3. (a) AUC. (b) F1-score.

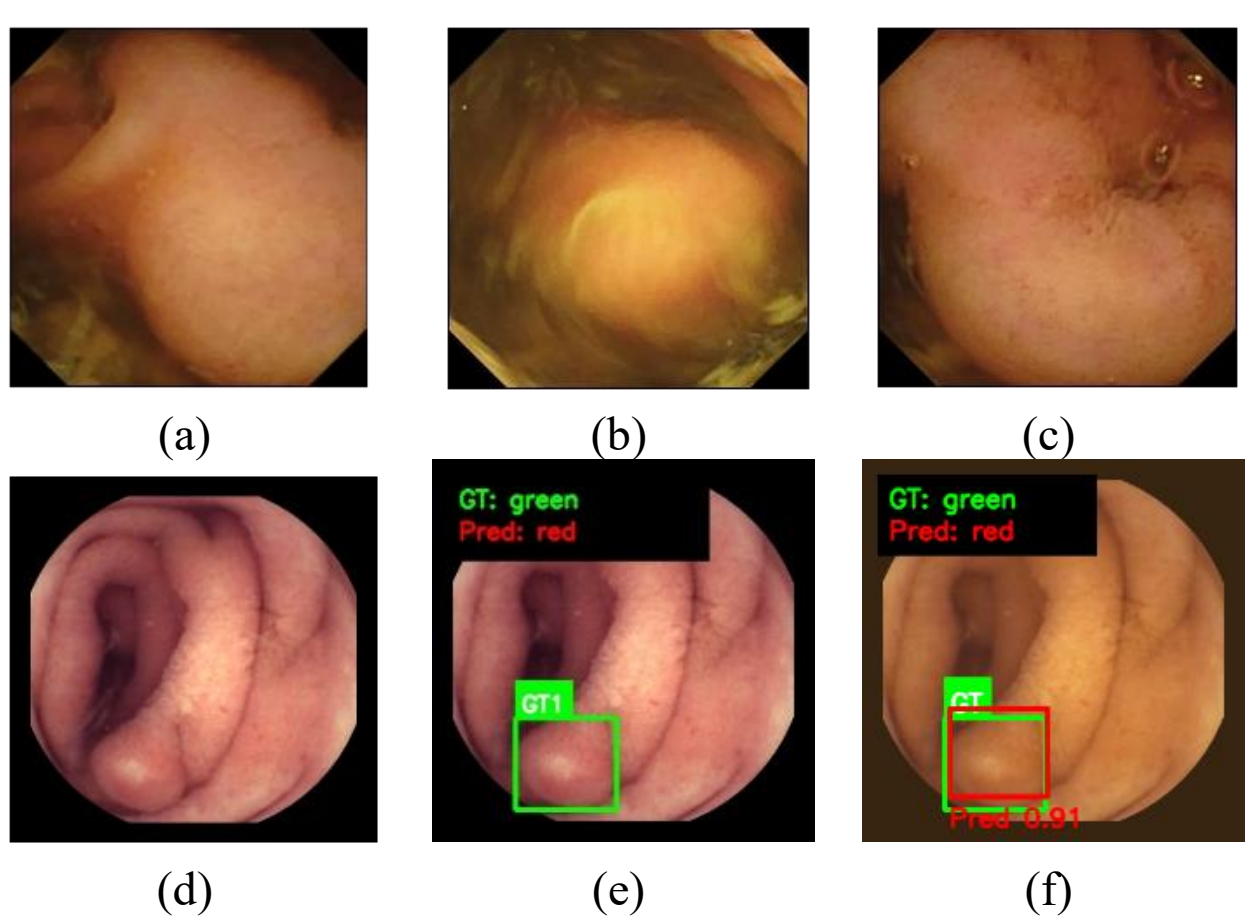

**Fig. 6**. (a–c) Reference images from DS4. (d) Raw image without transformation. (e) Failed polyp detection in the raw image. (f) Successful polyp detection in an image harmonized with the proposed method.

In terms of F1-score, the proposed method achieved the best results for ResNet50 and InceptionV3, reaching 82.9% and 86.0% (+3.7%), respectively. For YOLOv7, FDA produced the highest F1-score, but the proposed method still achieved a strong F1-score of 88.8% while also giving the highest AUC. The sensitivity and specificity results further support this trend: YOLOv7 improved from 77.2% sensitivity and 73.3% specificity in the raw setting to 90.9% (+17.7%) sensitivity and 80.0% (9.1%) specificity after harmonization, while InceptionV3 achieved 84.0% sensitivity and 83.3% specificity. These results indicate that the proposed method improves capsule endoscopy robustness by providing strong AUC, competitive or superior F1-score, and a more balanced operating behavior without requiring method-specific parameter tuning.

## 5. CONCLUSION

This paper presented a lightweight dataset harmonization step to mitigate cross-dataset appearance variation in endoscopy. Under a strict train-on-one, test-on-others protocol, harmonizing the target images consistently improved cross-dataset generalization for both detection (YOLOv7) and classification (ResNet50, InceptionV3), while keeping the same architectures, trained weights, and inference and evaluation settings unchanged. Qualitative examples further show that after harmonization, target frames better match the reference appearance and yield more reliable detections, particularly in challenging cross-domain cases. Compared with Z-score normalization, histogram matching, FDA, PCA/SVD color transfer, and PBCT, the proposed method achieved the best results in most dataset–backbone combinations and provided a more stable performance profile across AUC, F1-score, sensitivity, and specificity. Although the advanced baselines produced competitive results in some cases, they required method-specific parameter choices, such as spectral region size, color space, transformation strength, or palette size. In contrast, the proposed harmonization uses fixed reference-domain CIE-Lab statistics and does not require parameter tuning, making it simpler and more reproducible for cross-dataset endoscopic image analysis. Overall, the results indicate that simple appearance alignment can effectively reduce a practical source of generalization error and provide an effective preprocessing option when target datasets are small and retraining is undesirable.

Future work includes applying the proposed method as a client-side normalization step in federated learning to reduce inter-site appearance mismatch. Promising extensions also include other color medical imaging modalities, such as histopathology and retinal imaging.

## 6. ACKNOWLEDGMENT

This project has received funding from the European Union's HORIZON TMA MSCA Doctoral Networks programme (HORIZON-MSCA-2023-DN-01) under Grant Agreement No. 101169012 (INTELLI-INGEST Project).